\documentclass[graybox]{svmult}

\usepackage{type1cm}        
\usepackage{makeidx}         
\usepackage{graphicx}        

\usepackage{multicol}        
\usepackage[bottom]{footmisc}

\usepackage{newtxtext}       %
\usepackage[varvw]{newtxmath}       

\usepackage{color, soul}

\usepackage[natbibapa]{apacite}
\usepackage[bookmarks=true]{hyperref}
\usepackage[all]{hypcap}				
\usepackage{bookmark}
\definecolor{citationBlue}{RGB}{0,102,204}
\definecolor{linkRed}{RGB}{255,0,0}
\hypersetup{colorlinks=true,citecolor=citationBlue,linkcolor=linkRed,linktocpage=true}

\usepackage{listings}
\usepackage{caption}
\usepackage{indentfirst}
\usepackage{fancyhdr}
\usepackage{setspace}
\usepackage{amsmath,amsfonts}
\usepackage{breqn}
\usepackage{pgfgantt}
\usepackage{pdflscape}
\usepackage{float}
\usepackage{framed}
\usepackage{makecell}
\usepackage{soul}
\usepackage{bm}
\usepackage{multirow}

\usepackage[subnum]{cases}
\usepackage[toc,page]{appendix}
\usepackage{mathtools}

\usepackage[lined,boxed,ruled]{algorithm2e}

\usepackage{pgfplots}
\usepackage{tikz}
\usetikzlibrary{decorations.pathreplacing,
	backgrounds,
	fit,
	shapes.geometric,
	arrows.meta,
	petri,
	positioning}
\tikzset{%
	base/.style = {rectangle, rounded corners, draw=black,minimum width=2cm, minimum height=1cm,text centered, font=\sffamily},
	simulStarts/.style	= {base, fill=white!30},
	question/.style 	= {base, fill=blue!30},
	process/.style 		= {base, minimum width=2.5cm, fill=orange!15},
	dangerAct/.style 	= {base, fill=red!15, draw=red, font=\ttfamily},
	normalAct/.style 	= {base, fill=green!15, draw=green, font=\ttfamily},
	support/.style 		= {diamond, fill=blue!30, draw},
	point/.style		= {circle,radius=0cm,minimum size=0cm,inner sep=0cm, outer sep=0cm},
}

\DeclareMathOperator*{\argmax}{\textbf{argmax}}
\DeclareMathOperator*{\argmin}{\textbf{argmin}}

\newcommand{\tm}[1]{\text{#1}}

\makeindex             

\begin{document}
	
	\title*{Automated Driving Without Ethics: Meaning, Design and Real-World Implementation}
	\author{Katherine Evans, Nelson de Moura, Stéphane Chauvier, Raja Chatila}
	\institute{Katherine Evans \at IRCAI, Jožef Stefan Institute, Jamova cesta 39 SI-1000 Ljubljana, Slovenia; \email{katie.d.evans@gmail.com}
		\and Nelson de Moura \at INRIA, 2 Rue Simone Iff 75012 Paris, France; \email{nelson.demoura@inria.fr}
		\and Stéphane Chauvier \at Sciences, Normes, Démocratie, Sorbonne Université, 1 rue Victor Cousin 75005 Paris, France; \email{stephane.chauvier@sorbonne-universite.fr}
		\and Raja Chatila \at ISIR, Sorbonne Université, 4 Place Jussieu, 75005 Paris, France; \email{Raja.Chatila@sorbonne-universite.fr}}
	%
	%
	\maketitle
	
	\abstract*{The ethics of automated vehicles (AV) has received a great amount of attention in recent years, specifically in regard to their decisional policies in accident situations in which human harm is a likely consequence. After a discussion about the pertinence and cogency of the term ‘artificial moral agent’ to describe AVs that would accomplish these sorts of decisions, and starting from the assumption that human harm is unavoidable in some situations, a strategy for AV decision making is proposed using only pre-defined parameters to characterize the risk of possible accidents and also integrating the Ethical Valence Theory, which paints AV decision-making as a type of claim mitigation, into multiple possible decision rules to determine the most suitable action given the specific environment and decision context. The goal of this approach is not to define how moral theory requires vehicles to behave, but rather to provide a computational approach that is flexible enough to accommodate a number of human ‘moral positions’ concerning what morality demands and what road users may expect, offering an evaluation tool for the social perception of an automated vehicle’s decision making.}
	
	\abstract{The ethics of automated vehicles (AV) has received a great amount of attention in recent years, specifically in regard to their decisional policies in accident situations in which human harm is a likely consequence. After a discussion about the pertinence and cogency of the term ‘artificial moral agent’ to describe AVs that would accomplish these sorts of decisions, and starting from the assumption that human harm is unavoidable in some situations, a strategy for AV decision making is proposed using only pre-defined parameters to characterize the risk of possible accidents and also integrating the Ethical Valence Theory, which paints AV decision-making as a type of claim mitigation, into multiple possible decision rules to determine the most suitable action given the specific environment and decision context. The goal of this approach is not to define how moral theory requires vehicles to behave, but rather to provide a computational approach that is flexible enough to accommodate a number of human ‘moral positions’ concerning what morality demands and what road users may expect, offering an evaluation tool for the social acceptability of an automated vehicle’s decision making.\\ \\
	This is a preprint of the following chapter: Katherine Evans, Nelson de Moura, Stéphane Chauvier, Raja Chatila, Automated Driving Without Ethics: Meaning, Design and Real-World Implementation, published in Connected and Automated Vehicles: Integrating Engineering and Ethics, edited by Fabio Fossa, Federico Cheli, 2023, Springer Nature Switzerland AG reproduced with permission of Springer Nature Switzerland AG. The final authenticated version is available online at: \url{http://dx.doi.org/[insert DOI]}
	}
	
	\section*{Introduction}
	
	In recent years, the actions and decisions of artificial intelligent systems (AIS) have not only become a regular feature of many human social contexts, but have begun to threaten if not directly impact long-standing moral values such as human welfare, dignity, autonomy and trust. This capacity for topical and visceral moral impact has catapulted the field of machine ethics from its prospective and relatively experimental origins, into a new role as a critical body of literature and perspectives that must be consulted in the real-world design of many AIS. At this critical juncture, machine ethicists must therefore reassess many of the common assumptions of the field, not in the least what we might call the standard view of machine ethics, one which paints the route to morally acceptable behavior in machines as the product of an often complex simulation of human moral reasoning, moral theories, or substantive components of such capacities (i.e. consciousness, emotions, subjectivity) in these systems. Our aim in this chapter is not only to negate the viability of the simulated ethical decision-making that this approach prescribes, but to offer an alternative account that removes the need to place the seat of moral reasoning in the machines themselves. In effect, machines do not need to be robust artificial moral agents, nor moral partners of humans, in order to avoid making moral mistakes across their interactions with human agents. Rather, the original justification for moral decision-making in AIS–namely, that safety-related design concerns might include morally salient criteria as a result of the increasing automation of different aspects of the human social sphere–can be perhaps better satisfied by the creation of ethically-sensitive or minimally-harming machines that only borrow some structural aspects of the normative theories or decision procedures of human morality. 
	
	The first two sections of this chapter address misconceptions of machine morality, artificial morality, and a machine’s capacity for ethical decision-making directly, taking a strong deflationist stance on the proper course of machine ethics in application to real-world machines. The final section offers a positive account and approach of this deflationist attitude, introducing the Ethical Valence Theory, alongside an application case of ethical decision-making in automated vehicles. It also tangentially addresses an important distinction in the state of the art of machine ethics, between what we will call structural ethics–known to many as AI principles, such as accountability, privacy or transparency– which focus on the compliance of machines and their larger sociotechnical systems with certain principles or moral standards; and tactical ethics, which aims to ensure that the decisional output of machines aligns with the moral expectations of its end-user(s), or society at large.

	\section{The Semantics Of Automated Driving}
	\label{sec:1}
	
	Before broaching  the discussion concerning  whether an automated vehicle (AV) might be endowed with ethically-aware decision-making, it is necessary to clarify some misconceptions about the goals and limitations of machine ethics given the current technological state of the art, and how such ideas can be implemented when applied to the AV domain. The first two sections of this article address these misconceptions directly, and the final section introduces a positive account of ethically-aware decision-making in automated vehicles. 
	
	\subsection{The Problem with Machine Ethics}
	\label{subsec:1.1}
	
	There exists a philosophical stance concerning machine ethics which has precipitated a whole stream of research and publications, one which considers machines endowed with artificial intelligence capacities as being able to make deliberate moral decisions pertaining to what is right or wrong, good or evil. In the media and among the general public, this has translated into the belief that such machines do indeed have moral agency. A logical consequence of this attitude is to consider that such machines are responsible and accountable for their actions. A further consequence is to claim that they should be endowed with a legal, if not a social form of personhood, possibly with related rights and duties.
	
	Approaches to "artificial morality" are based on formalizing values, and on translating deontic or consequentialist or other moral theories into mere algorithms, and on proceeding to apply formal logic reasoning or function optimization to make decisions.
	
	These systems implement automated decision-making. They are sometimes dubbed as “autonomous” (another confusing concept that we will not address here), when in a given domain and for given tasks, they are capable of accomplishing the specified tasks on their own despite some changes in this domain’s environment. While the above briefly summarizes a typical approach to implementing ethical decision-making in machines, we maintain that this approach is flawed. Indeed, as we will argue next, these systems do not perform any ethical deliberation.
	
	By definition, a computer can only run algorithms. A computational intelligent system is a set of algorithms using data to solve more or less complex problems in more or less complex situations. The system might include the capability of improving its performance based on learning, the currently dominant stream in Artificial Intelligence. Learning - be it supervised or unsupervised - is a computational process using a set of algorithms that classify data features to provide statistical models from the data. These models are then used to predict future situation outcomes and to make decisions consequently. Reinforcement  learning is a method to sequentially optimize an objective function, which is a mathematical expression of a reward function, evaluating previous decisions to achieve more efficient ones, i.e., those increasing the total expected reward.
	
	Machines operate at the computational syntactic level. They have no understanding of the real-world meaning of the expressions, the formulae or the numerical values they manipulate. Semantics in their representations are provided by programmers and users. An image labeled “cat” in a dataset doesn’t provide any knowledge to the machine about what a real cat is. A statistical model built by machine learning, or a model using hand-made semantic descriptors, still do not carry any meaning about reality to the system. This is why we affirm that machines actually have no semantics. In addition, machines have no knowledge of the reasons behind what mattered for the development of the algorithms and the design choices made by programmers. In systems based on statistical machine learning, there is no causal link between inputs and outputs, merely correlations between data features.
	
	Fundamentally, algorithms and therefore machines lack semantics and contextual awareness.
	
	Human values such as dignity or justice (fairness, equality, ...) are abstract and complex concepts. Dignity, for example, which is intrinsically attached to human beings, is an intricate concept and is not computable. It cannot be explicitly described or taught to machines. The respect of such values as a basis for human action is not reducible to simple deductive computations, or to statistics about previous situations. Human ethical deliberation is grounded in history, in lifelong education and in reflections about these concepts, about a substantive view of the "good life", society, and the actual context of action. Understanding a context is in itself not reducible to a set of numerical parameters. A context takes into account the past, the evolving present and the future. A situation for moral deliberation has a spatial and temporal dimension, and uses abstract notions related to the societal context, all of which are absent from the data fed to machines and from their algorithmic operations, and a fortiori cannot even be described meaningfully in these terms. 
	
	Thus, machines cannot determine ethical values and cannot make ethical decisions, but this does not preclude them from performing actions that have ethical impacts from the perspective of humans. This perspective stands in sharp contrast to earlier espousals of artificial morality, which often point to the need for relatively robust simulations of human moral reasoning in machines \citep{wallach2009moral}, or more broadly, that instilling general ethical principles in machines, and thus developing artificial moral agents, is both possible and necessary for successful human-machine interaction \citep{anderson2010robot}. van Wynsberghe and Robbins \citep{vanWynsberghe2019critiquing} rightly criticize these views. We position ourselves closer to Fossa’s “Discontinuity Approach” \citep{fossa2018artificial}.  
	
	Humans tend to naturally, but wrongly, project moral agency onto machines. Our approach is to dispossess machines of this agency, to make it clear that morality is in this case only a human conception and interpretation of a machine action which has no moral dimension.

	\subsection{Addressing the AV's Problem}
	\label{subsec:1.2}
	
	If we adopt the metaphysically deflationist attitude that we have just delineated and reject the mythology of artificial moral agency, does it mean that we also have to give up any idea of machine ethics? Is machine ethics a simple by-product of the mythology of artificial moral agency? Of course not, since even if we cease to view an AV as an exotic mind located in a coach-building, an AV is still much more than a technological tool whose ethical regulation falls entirely on the human designer/provider/user side. If there is something mythological in the idea of autonomous artificial agent or of artificial autonomy, even if there is no such thing as self-purposing agency in a machine, there is still a crucial difference between a machine that delivers its effect as a response to a command, even if the machine can calibrate the effects it delivers according to the context of their delivery (auto-regulative machines), and a machine that can act spontaneously in the social context in which it is designed to deliver its effects. Without entering into the classical metaphysical debate regarding human free will, the refusal of any capacity of self-determination to a creature does not imply the refusal of any capacity to spontaneous acting. Spontaneity does not mean self-determined acting, but self-originated acting. Let’s illustrate this idea by imagining a spontaneous coffee-delivery machine. It does not deliver coffee as a response to a command (the introducing of a coin), but it delivers coffee to whomever its program instructs it to. The in-want-of-coffee consumer has to stand in the front of the machine, he has to look at it intensely and perhaps the machine will deliver him a cup of coffee, perhaps not. He does not know why, but it is the way the machine functions. Of course, there is nothing mysterious inside the machine, no soul, no mind, no free will. There is only an algorithm that realizes a calculus whose result is that sometimes the machine delivers a coffee, sometimes not. It is obvious that we are here in what we could call the circumstances of ethics. For suppose that we discover, by observing the machine, that it spontaneously delivers coffee to young men but never to old women. Of course, the machine does not have any notion of gender and age differences, no philosophy of such differences, no value judgments concerning men and women, young and old persons, but if it is designed to spontaneously deliver effects that are meaningless to the machine, these effects have an important meaning for us, they can be viewed as an instance of discriminatory action.
	
	This small imaginary case aims to illustrate that a specific ethical problem arises from what we can label spontaneously acting machines (SAM): because these machines spontaneously deliver effects that can have an ethical resonance for the human addressees of these effects, we cannot implement these machines into the social world without including into the program of functioning of these machines an algorithmic regulation that constrains the range of effects that it can spontaneously deliver. We have to make the machine a never-harming machine or, at least, a minimally-harming machine. The question is not to give to the machine a reflexive capacity to morally evaluate its operations. Machine ethics is not the ethics of the machine, the ethics that the machine itself follows. Machine ethics is the non-harming behavior that we have to impose on the SAM and whose technical translation lies in including into the program - or what some might consider the “soul” of the machine - a way of calculating its optimal effects which guarantees that these effects will be, for us, minimally harming or, even preferably, harm-free.
	
	This is the core idea of the Ethical Valence Theory (EVT) that applies this general idea to the regulation of the social behavior of an AV. EVT does not aim to make an AV a moral partner of human drivers, as if, in a not-too-distant future, we will drive surrounded by genuinely Kantian or Millian cars. It only means that the program of optimization that is the “soul” of the AV includes a set of restrictions and of mitigation rules that is inspired by Kantian or Millian ethics and will guarantee that the operations of the car (braking, accelerating, etc.) will be viewed by us as minimally harming when effectuated in a social context including others vehicles, passengers, pedestrians, etc.

	\section{Implementing Ethically-Aware Decision-Making}
	\label{sec:2}				
	
	It has been established in the previous section that machines that act in the social world cannot be deployed without being able to mitigate the moral consequences of their actions and minimize or eliminate the risk of harm for human agents. This section sets the scope for these decisions (subsection \ref{subsec:2.1}), a proposition of claim mitigation for dilemma situations using the ethical valence theory (EVT, subsection \ref{subsec:2.2}) for the non-supervised driving task and its implementation (subsections \ref{subsec:2.3} and \ref{subsec:2.4}), considering a generic AV architecture.
	
	\subsection{The Scope of Ethical Deliberation}
	\label{subsec:2.1}
	
	Initially, the deflationism implied in the conceptual shift from artificial moral agent to simple spontaneously acting machine (SAM) may appear to leave no room for machine ethics, or a forteriori moral theory, in the design of automated vehicles. Instead, the design goal of a minimally-harming machine might rather fall under the purview of product safety \citep{cawthorne2020ethical}, and more precisely still, the minimization of physical harm to any human agent with whom the AV interacts. Under this view, if ethics applies at all, it does so only when evaluating certain structural design choices in the larger sociotechnical system of which physical automated vehicles are a relatively small part: we may, for instance, find it unethical that an AV manufacturer fails to disclose the full breadth (or limitations) of an AV’s functions to its end user, or that certain AV service providers collect road user data for applications that are opaque, vaguely defined, or if this data is used to infer new facts about these users without their consent. These ethical concerns are typically captured by so-called ‘ethical design principles’ such as accountability, transparency, responsibility or privacy \citep{ieee2017ethically, dignum2019responsible}, and have played a central role in the current normative landscape of intelligent system design. Let us call this particular application of machine ethics {\it structural}\footnote{Among the now numerous ontologies of machine ethics, our concept of structural machine ethics coincides with, but is not limited to ‘Ethics of Designers’ \citep{dignum2019responsible}, since it focuses exclusively on the actions of human agents in sociotechnical systems, but expands this class of scrutable agents to include other stakeholders such as service providers and regulators.}, since it yields normative criteria that impact the design structure of the technical artifacts themselves, but also the collaborative structure of the different human stakeholders involved in the artifact’s sociotechnical system.
	
	By the lights of these distinctions and those made in section \ref{subsec:1.2}, it would then seem that automated vehicles are wholly the concern of structural ethics in regular driving conditions. This is so because, in contradistinction to our biased coffee machine example, the automated vehicle does not make critical discriminatory decisions which impact multiple human agents during normal function within its operational design domains (ODDs). It is possible, in other words, for the vehicle to successfully harm no one across its tactical planning, since the claims to safety of all road users are jointly satisfiable by the vehicle. In ODDs such as an open highway or a private industrial road, this would imply that the vehicle need only satisfy the AV occupant’s claim to safety, while in more complex ODDs such as peri-urban driving, this might imply observing additional safety standards, or constraints to respect alongside the satisfaction of the occupant’s claim, such as ensuring longer following distances or lower speeds when overcoming vulnerable road users such as cyclists. In the terms of our biased coffee machine example, this would be akin to satisfying the claims every customer has to coffee. Structural machine ethics allows us to appraise and eventually improve the degree to which the machine treats users, and especially end-users in practice, ensuring that it does so transparently, responsibly and in respect of the user’s privacy, but its purview does not extend past these brightlines.
	
	Consider now, however, what structural ethics recommends in situations where automated vehicles must contend with multiple claims? Importantly, this does not exclusively imply that the machine must consider the welfare of multiple end users, it also intriguingly covers the more common case of a machine which, in pursuit of the welfare of one human agent, may indirectly cause harm to another in ways which appear ethically salient to this second user, other human agents, or society at large. More succinctly, harm to human agents other than the end user is an external effect of the pursuit of the end user’s safety, but each agent holds a moral expectation as to how a balance should be struck. Rather lyrically, this might occur when our coffee machine must decide whether to drop its last remaining dose of sugar into the cup of its very thirsty end user, or into that of a person with dangerously low blood sugar levels. More pragmatically, this could occur in abnormal driving conditions in automated vehicles, such as when an inevitable collision or risky maneuver pits passenger safety against that of any other road user. In both cases, beyond trivially recommending that the resultant decisions of both machines be transparent, responsible, accountable and private where applicable, structural ethics fails to recommend a solution as to how to mitigate these competing claims; it cannot itself prescribe how to distribute safety or sugar in situations where not everyone can receive a share.
	
	Instead, the clear ethical salience of these latter types of cases suggests the need for a second and conceptually separate application of ethics, what we will call {\it tactical} ethics. Here, the goal is not to ensure that the design structure of intelligent system design beget accountable or privacy-respecting systems, but rather that the resultant decisions of these systems align with the moral expectations of the individuals which they affect. Tactical ethics provides normative criteria and decision procedures which impact or constrain the tactical planning of a given machine in contexts of ethical salience. In such situations, these criteria allow the machine to recognize and minimize the moral harm brought about by the pursuit of its practical goals, and to mitigate the competing moral claims individuals hold over the behavior of the machine when not all claims can be jointly satisfied. Tactical ethics may recommend, for instance, that an automated coffee machine operate on a ‘first come, first serve’ basis when distributing limited resources to multiple customers, or that it not distribute caffeinated drinks to children, and never more than 6 cups per day per customer. In the case of automated vehicles, tactical ethics might recommend that the safety of child pedestrians should always be prioritized against passenger safety or comfort, or that the vehicle should always privilege the safety of the most vulnerable road user during unavoidable collisions.
	
	In sum, tactical ethics ensures that machines behave as if they were moral agents in situations where moral agency is likely required or expected in the eyes of society. Rather than creating artificial moral agents which simulate human processes of ethical reasoning and judgment to achieve this end, tactical ethics achieves ethical responsiveness in machine behavior via a process of moral contextualization of the machine’s operational design domain. During the design process, human experts first define the relevant marks of ethical salience which occur naturally in the ODD. These are facts or features of a given design context which may carry ethical weight, such as the fact that a given individual is the AV’s passenger, or that caffeine consumption can be harmful to children. Given these marks of ethical salience, human experts then devise what we can call {\it profiles}: various decision procedures that take these moral facts into account and provide different methods for mitigating the degree of the machine’s responsiveness to each claim in its environment: for instance, ‘first come, first serve for caffeinated drinks, unless the customer is a child’, or ‘privilege passenger safety in unavoidable collisions, unless this action is lethal to another road user’.
	
	While the ontological structure, axiology or founding principles of these different profiles may beneficially take inspiration from established moral theories such as utilitarianism, contractarianism or Kantian ethics, these profiles need not robustly or exhaustively track the types of decisional processes or resulting normative recommendations these theories typically yield. Instead, this process of moral contextualization aims at generating profiles which track common sense morality, the content of which is ideally informed by descriptive ethics, or empirical studies into acceptability. The next section briefly introduces one such approach to tactical ethics, the Ethical Valence Theory (EVT), contextualized for unavoidable collision scenarios in automated vehicle driving.

	\subsection{Ethical Valence Theory}
	\label{subsec:2.2}
	
	The approach behind the Ethical Valence Theory is best understood as a form of moral claim mitigation, the fundamental assumption being that any and every road user in the vehicle’s environment holds a certain claim on the vehicle’s behavior, as a condition of their existence in the decision context. Conceptually, the EVT paints automated vehicles as a form of ecological creature \citep{gibson1979theory,gibson2022theoretical}, whose agency is directly influenced by the claims of its environment. Given the particulars of a decision context, these claims can vary in strength: a pedestrian’s claim to safety may be stronger than that of the passenger’s if the former is liable to be more seriously injured as a result of an impact with the AV\footnote{Analytically, individual claims can be understood as contributory or pro tanto reasons for the vehicle’s acting a certain way \citep{dancy2004ethics,prichard2002basis}.}. The goal of the AV is to maximally satisfy as many claims as possible as it moves through its environment, responding proportionally to the strength of each claim. When a conflict across claimants occurs, the vehicle will adopt a specific mitigation strategy (or profile) to decide how to respond to these claims, and which claims to privilege.
	
	Within the structure of the Ethical Valence Theory, the role of claim mitigation is to capture the contribution that moral theory could make to automated vehicle decision-making. Claims, in other words, allow the vehicle to ascertain what morality requires in ethically salient contexts, by tracking how fluctuations in individual welfare affect the rightness or wrongness of an AV’s actions. Specifically, in some respects, the EVT takes analytical inspiration from both the ‘competing claims’ model popular in distributive ethics \citep{nagel2012moral,voorhoeve2014should}, and Scanlonian contractualism \citep{scanlon1998we}. These theories, far from providing the objectively ‘right’ answer to AV ethics, provide fruitful starting points for tactical ethics in automated vehicles for one simple reason: they afford multiple normatively relevant factors (such as agent-relative constraints and options \citep{kagan1992limits}) to play a role in what matters morally. Since the goal of tactical ethics is to track and eventually satisfy public moral expectations as to AV behavior, it is important that the use of moral theory not frustrate this aim by revising upon common sense morality, or more specifically, the kinds of factors or features that can ground claims. In this sense, if a passenger feels that she holds a certain claim to moral partiality over the actions of the vehicle, for instance in virtue of the fact that she owns or has rented the AV, or views it as a type of moral proxy for her actions \citep{keeling2019trolley,johnson2008computers} it is important that moral partiality feature in the marks of ethical salience originally identified by human experts, and thus in the resultant claim mitigation strategy. While some moral theories such as contractualism naturally accommodate these sorts of factors, viewing them as legitimate sources of moral obligation, others such as utilitarianism, do not. For a more exhaustive discussion of the interaction between the EVT and moral theory, see \citep{evans2020ethical}.
	
	Importantly, the correlation between the recommendations of even contractualist-type moral theories and public expectations as to AV behavior only go so far. To this end, and in order to take public expectations seriously, moral theory must be complemented by empirical accounts of acceptability. Within the structure of the Ethical Valence Theory, this is accomplished through the concept of a valence: a weight added to the strength of each individual’s claim in the vehicle’s environment, which fluctuates in relation to how that individual’s identity corresponds to set of acceptability-specific criteria, such as different age groups, genders or road-user types. These criteria are then organized into categories, and eventually hierarchies, delineating various strengths of valences. Here, we must be careful not to confuse these sorts of acceptability criteria with simple societal biases, and thereby view a valence as something which represents a given individual’s ‘popularity’ in society (as was arguably the case within the Moral Machine Experiment \citep{bonnefon2016social}). The aim of valences is not to thwart important moral principles such as equality or impartiality by picking out the darlings of a society in the AV’s decision context. Rather, it is to ensure that important normative factors and features (such as relative user vulnerability, fairness or even liability) effectively impact AV decision-making, tipping the scale so-to-speak, in situations where the assessment of physical harm alone may not result in an acceptable decision on the part of the AV. Put another way, it is only through the inclusion of valences in claim mitigation that we arrive at decision recommendations with that casuistic flavor so characteristic of common-sense morality, such as ‘prioritize the safety of the most vulnerable road user, unless this action results in the death of a child’.
	
	The final conceptual piece of the Ethical Valence Theory is the notion of a moral ‘profile’: a specific decision procedure or method which mitigates the different claims and valences of road users. Essentially, each moral profile provides a specific criterion of rightness: a maxim or rule which decides the rightness or wrongness of action options. In this way, a moral profile also dictates which claims the AV is sensitive to and when, and how those claims are affected by a given individual’s valence strength. While there are surely many ways to organize the mitigation process between valences and claims, one way to reflect the aforementioned concept of harm as an external effect of the pursuit of end-user safety is to make a preliminary categorial separation between those users inside the AV and those outside of it, thereby painting claim mitigation as the balancing of the passenger(s) claim versus those in the AV’s surrounding environment. Then, there are a number of potential mitigations we could find across these two interest groups: a risk-averse altruist moral profile would, for example, privilege that user who has the highest valence in the event of a collision, so long as the risk to the AV’s passenger is not severe. Intuitively, this type of profile supports the view that the passenger of the AV may be willing to incur some degree of harm in order to respond to the claims of other users in the traffic environment, but not so much that he will die or suffer seriously debilitating injuries as a result. In the end, there is no single profile that will once and for all resolve the moral and social dilemma of automated vehicle behavior. Instead, the choice of moral profile, along with the choice of valence criteria, exist as different entry points for human control in automated decision-making. Importantly, these profiles, claims and valences need not be intemporal, universal or unilateral. Rather, they can and likely ought to be continuously reviewed and updated, as fluctuations in AV adoption, public acceptability and legal constraints evolve within each operational design domain.

	\subsection{The Technical Implementation of Ethical Decision-Making}
	\label{subsec:2.3}
	
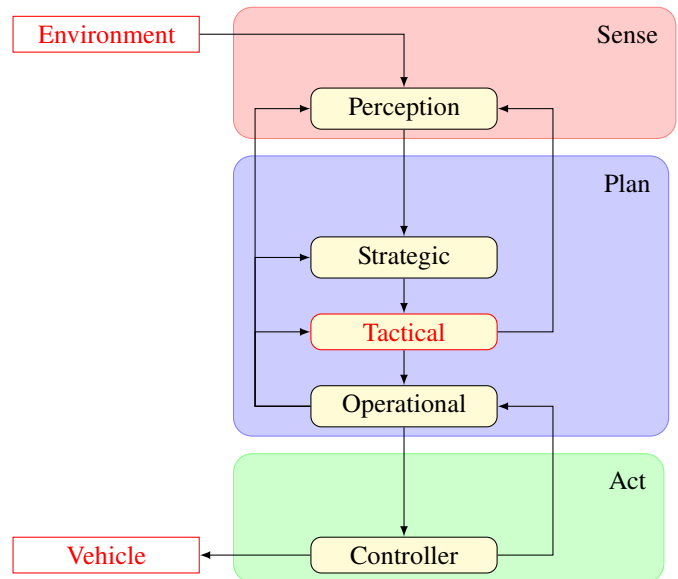
\begin{figure}[!ht]
		\begin{center}
			\resizebox{90mm}{!}{
				\begin{tikzpicture}
					
					\tikzstyle{inout}	= [rectangle,draw=red,fill=white!50,text=red, minimum width=2.5cm]
					\tikzstyle{back}	= [rounded corners=0.25cm]
					\tikzstyle{box}		= [rectangle, draw=black, fill=yellow!20, text=black, minimum width=2.5cm, rounded corners]
					\tikzstyle{main}	= [rectangle, draw=red, fill=yellow!20, text=red, minimum width=2.5cm, rounded corners]
					\tikzstyle{dummy}	= [circle, radius=0cm]
					
					\tikzstyle{down}	= [->,>=latex]
					
					\node[inout]	(envi)	at (-4,7)	{Environment};
					
					\node[]			(sens) 	at (3,7)	{Sense};
					\node[box] 		(perc) 	at (0,6) 	{Perception};
					\node[dummy]	(dum1)  at (-2,6)	{};
					
					\node[]			(plan) 	at (3,5)	{Plan};
					\node[box] 		(stra) 	at (0,4) 	{Strategic};
					\node[main]		(tact) 	at (0,3) 	{Tactical};
					\node[box] 		(oper) 	at (0,2) 	{Operational};
					\node[dummy]	(dum2)  at (-2,2)	{};
					
					\node[]			(act) 	at (3,1)	{Act};
					\node[box] 		(cont) 	at (0,0) 	{Controller};
					\node[inout] 	(vehi) 	at (-4,0)	{Vehicle};
					\node[dummy]	(dum3)  at (-2,0)	{};
					
					\draw[down]	(perc) -- (stra);
					\draw[down] (stra) -- (tact);
					\draw[down] (tact) -- (oper);
					\draw[down] (oper) -- (cont);
					\draw[down] (cont) -- (vehi);
					\draw[down] (envi) -- (0,7) -- (perc);
					
					\draw[down] (oper) -- (-2,2) -- (-2,3) -- (tact);
					\draw[down] (oper) -- (-2,2) -- (-2,4) -- (stra);
					\draw[down] (oper) -- (-2,2) -- (-2,6) -- (perc);
					
					\draw[down] (tact) -- (2,3) -- (2,6) -- (perc);
					
					\draw[down] (cont) -- (2,0) -- (2,2) -- (oper);
					
					\begin{scope}[on background layer]
						
						\node[back, fill=red!20, draw=red!50, fit=(perc) (sens) (dum1)] {};
						\node[back, fill=blue!20, draw=blue!50, fit=(plan) (stra) (tact) (oper) (dum2)] {};
						\node[back, fill=green!20, draw=green!50, fit=(act) (cont) (dum3)] {};
						
					\end{scope}
					
				\end{tikzpicture}
			}	
		\end{center}
		\caption{AV's decision-making structure used in \citep{thrun2006stanley}}
		\label{fig:1}
	\end{figure}

	The human driver decision-making process under normal conditions can be divided into three phases \citep{michon1985critical}: strategic, tactical and operational (represented by the Plan block of figure \ref{fig:1}). This structure is used in most implementations of decision-making algorithms for AV's (except in end-to-end planning \citep{schwarting2018planning}) and its use in real prototypes dates back to \citep{thrun2006stanley}. It is the tactical layer that holds the responsibility of accounting for the behavior of local road users and thus, dealing with any dangerous situations that may arise which impose some risk distribution concerning the AV's actions towards the other users in the environment.
	
	A plethora of different algorithms have been used to implement the tactical layer, from the common Markovian approaches \citep{ulbrich2013probabilistic, demoura2020ethical, pusse2019hybrid}, end-to-end learning \citep{jaritz2018end} or more low-level methods such as RRT \citep{feraco2020local} and fuzzy approaches \citep{claussmann2018multi}. The work done in \citep{demoura2020ethical} will be used as a typical decision-making algorithm and modified in both deliberation under normal situations and deliberation in dilemma scenarios to adhere to the approach presented in this publication to treat the AV's problem.
	
	In the original version of the Markov decision process (MDP) proposed in \citep{demoura2020ethical}, during the evaluation of rewards for each state, according to the action considered and the next state predicted, all actions that resulted in a collision were removed from the viable action set for the policy optimization but left as an option during the evaluation of the Bellman's equation (\ref{eq:1}). The cost (negative reward) was fixed and equal for every collision, and therefore resulted in an equal degree of repulsion to the collision state independently of the road user involved.  
	
	\begin{equation}
		\label{eq:1}
		\begin{aligned}
			V_{t+1}(s_t) = \max_{a\in A}\biggl[R(s_t,a,s_{t+1})+\biggr. 
			\biggl.\gamma\cdot\sum_{s_{t+1}}P(s_{t+1}|s_t,a)V_{t}(s_{t+1})\biggr]
		\end{aligned}
	\end{equation}
	
	This procedure originates from the consideration that in normal situations, the decisional process should not be polluted by the ethical evaluation of the current environment configuration. However, as said in section \ref{sec:1}, clear constraints on the risk that the AV's actions might bring to other road users are needed, in all situations. Equation \ref{eq:2} represents this previously defined reward at a state $s_t$ given that an action $a_t$ will be executed, leading the AV to the state $s_{t+1}$:
	
	\begin{align}
		\label{eq:2}
		R(s_t,a_t,s_{t+1})=
		\begin{cases}
			s_{\tm{perf}}+s_{\tm{conseq}} &\tm{, if there are no collisions} \\
			c_{\tm{col}} &\tm{, otherwise}
		\end{cases}
	\end{align}
	
	The term $s_{\tm{conseq}}$ is composed by a cost\footnote{Cost and reward are used interchangeably to represent the quantity (negative or positive, respectively) which will determine the AV's optimal value.} related to adherence to the traffic code and another related to the physical proximity between the AV and other road users, evaluated by equation \ref{eq:3}. This quantity measures the risk of the action to other road users given the AV's current state, by evaluating how much the AV's trajectory will be convergent to the predicted trajectory of others (assuming that proximity means an increase of risk of injury). Again, the cost calculated by equation \ref{eq:3} (always negative) does not account for the nature of the road user in question while the term $s_{\tm{conseq}}$ represents the sum of each $s_{\tm{prox}}$ calculated comparing each possible AV-road user pair. 
	
	\begin{equation}
		\label{eq:3}
		s_{\tm{prox}} = c_{\tm{st}} + w_v\cdot\left[ \Delta v_{t+1}^{\tm{proj}} - \Delta v_{t}^{\tm{proj}} \right]
	\end{equation}
	
	Given that the EVT considers that each road user has a claim for safety that might vary in intensity, it should be reflected in the risk measurement as another parameter to the cost definition. As such, the $s_{\tm{conseq}}$ term would be given by equation \ref{eq:4}, where the weight $w_{\tm{ETV}_i}$ represents the strength relationship of the ethical valence from the AV (its passengers) and the other road user being considered. Such a parameter does not change with time or with any other property, it is only dependent on the valence of each road user being considered. The same parameter would be valid for the cost given to a collision, formulating the new reward function as equation \ref{eq:5} shows. 
	
	\begin{equation}
		\label{eq:4}
		s_{\tm{conseq}} = s_{\tm{traf}} + \sum_{i=1}^{i<N_{\tm{ag}}}w_{\tm{ETV}_i}\cdot s_{\tm{prox}}^{\tm{av,ru}_i}
	\end{equation}
	
	\begin{align}
		\label{eq:5}
		R(s_t,a_t,s_{t+1})=
		\begin{cases}
			s_{\tm{perf}}+s_{\tm{conseq}} &\tm{, if there are no collisions} \\
			w_{\tm{ETV}_i}\cdot c_{\tm{col}} &\tm{, otherwise}
		\end{cases}
	\end{align}
	
	The use of the weight $w_{\tm{ETV}_i}$ would therefore express, during the reward evaluation, a risk distribution that is compatible with the hierarchy of safety claims defined by the EVT, in all driving situations, normal or in a collision. For example, when the AV is approaching some road user that has a claim that is more important than the AV passengers, the weight should increase so as to push the AV to put some distance between both or to decelerate; thus another state with smaller velocity or a different position would conserve a higher reward in comparison with the first one. It should decrease, for the AV, in the opposite case as well, when the safety claim of the AV passenger is higher than the other road user(s).
	
	In the equation that defines the reward (equation \ref{eq:5}, with $s_{\tm{perf}}$ given by \ref{eq:6}, where $q_{\tm{lat}}$ refers to $d_{\tm{lat}}$, $q_{\tm{eta}}$ refers to $d_{\tm{obj}}$ and $v_{\tm{av}}^{\tm{proj}(\hat{d}_{\tm{obj}})}$) for each state there are multiple parameters that need to be defined, in the form of weights and thresholds. A machine learning approach can be used to estimate the values of these parameters using a set of training scenarios given by one of the many driving datasets available today, while maintaining the explainability and understandability property \citep{barredo2020explainable} of the algorithm since the reward function is predefined.
	
	As for the procedure to estimate the ethically-related parameters\footnote{Which does not include the ethical valence, that is set according to strength of the claim of each road user and must be defined according to societal and acceptability concerns, so as to reflect a population's "common sense" related to ethical dilemmas.}, a first fitting of the expression that represents $w_{\tm{ETV}_i}$, which can be a set of constant relations, polynomial or any other kind of function, must be done before the other parameters with specially selected dilemma situations. Such a procedure might be constructed from the ground up using standards that can dictate which type of collisions and in which road environments should be allowed, forbidden or even considered when this weight function is calculated.
	
	\begin{equation}
		\label{eq:6}
		s_{\tm{perf}} = w_{\tm{lat}}\cdot q_{\tm{lat}} + w_{\tm{dir}}\cdot\Delta\theta + w_{\tm{eta}}\cdot q_{\tm{eta}}
	\end{equation}
	
	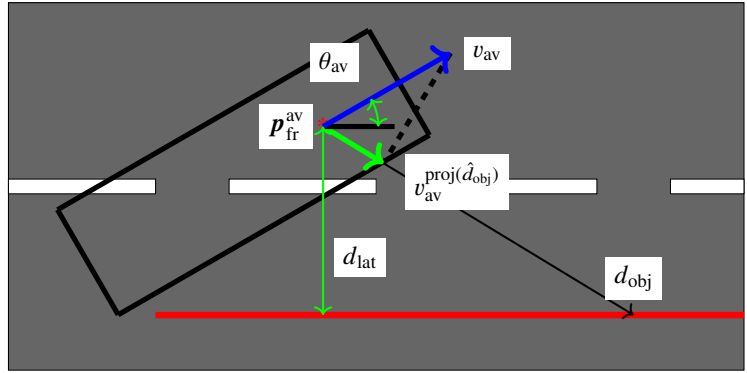
\begin{figure}[!ht] 
		\begin{center}
			\resizebox{100mm}{!}{
				\begin{tikzpicture}
					
					\tikzstyle{comp}=[rectangle,draw,fill=yellow!50,text=black]
					\tikzstyle{flux}=[->,very thick,>=latex]
					
					\tikzstyle{traj}=[line width=2.5pt]
					\tikzstyle{marking}=[rectangle,draw,fill=white,text=black]
					\tikzstyle{road}=[rectangle,draw,fill=black!60,text=black]
					\tikzstyle{sidewalk}=[rectangle,draw,fill=black!20,text=black]
					
					\tikzstyle{av}=[draw,rotate=0,line width=2pt]
					
					\draw[road] (0,2.5) rectangle (10,7.5);
					\draw[marking] (0,4.9) rectangle (2,5.1);
					\draw[marking] (3,4.9) rectangle (5,5.1);
					\draw[marking] (6,4.9) rectangle (8,5.1);
					\draw[marking] (9,4.9) rectangle (10,5.1);
					\draw[traj,red]	(2,3.25) -- (10,3.25);
					
					\node[fill=white!30, xshift=-5mm] (front axis)	at	(4.2775,5.815) {$\bm{p}_{\tm{fr}}^{\tm{av}}$};
					
					\draw[av] (4.898,7.126) -- (5.723,5.697);
					\draw[av] (5.723,5.697) -- (1.492,3.254);
					\draw[av] (1.492,3.254) -- (0.667,4.683);
					\draw[av] (0.667,4.683) -- (4.898,7.126);
					
					\draw[->,thick] (4.2775,5.815) -- (8.5,3.25) node[fill=white!30,yshift=5mm] {$d_{\tm{obj}}$};							
					\draw[av,dashed] (6.0096,6.8150) -- (5.099,5.3160);																		
					\draw[->,line width=2pt,green] (4.2775,5.815) -- (5.099,5.3160) node[black,fill=white!30,yshift=-2.5mm,xshift=10mm] {$v_{\tm{av}}^{\tm{proj}(\hat{d}_{\tm{obj}})}$};	
					\node[text=red] 					(front point)	at	(4.2775,5.815) {*};
					
					\draw[av] (4.2775,5.815) -- (5.25,5.815);
					\draw[->,line width=2pt,blue] (4.2775,5.815) -- (6.0096,6.8150) node[black,fill=white!30,yshift=0mm,xshift=5mm] {$v_{\tm{av}}$};
					
					\draw[<->,thick,green] (5.0275,5.815) arc [start angle=0, end angle=30, radius=0.75] node[black,yshift=5mm,xshift=-5mm,fill=white!30] {$\theta_{\tm{av}}$};
					
					\draw[<->,thick,green] (4.2775,5.815) -- (4.2775,3.25) node[black,midway,fill=white!30,xshift=5mm,yshift=-5mm]{$d_{\tm{lat}}$};
					
				\end{tikzpicture}			
			}
		\end{center}
		\caption{Performance reward parameters (from \citep{moura2021governing})}
		\label{fig:2}
	\end{figure}
	
	The fitting procedure for the non-ethical parameters can be done in iteration with the ethical parameters, so as to direct the convergence of the fitting for a domain where the ethical constraints are respected. This training structure avoids situations where the desired effects proposed by the ethical constraints are invalidated by the accommodation of other non-ethical parameters.
	
	However, in some cases the dilemma situation is inevitable since urban environments are highly dynamic and a single unpredictable action from a road user can create a cascade of consequences that is beyond the hypothetical capability of a future AV to account for, as explained previously and in \citep{moura2021governing}. In such cases, where the AV predicts that all potential actions lead to a collision, to consider a performance criteria or even the proximity one modified in this subsection does not make sense. In this specific situation the expected harm that the AV will inflict on others and its passengers needs to be taken into consideration for the action deliberation, together with the ethical valence of each of the road users. 
	
	\subsection{Deliberation in Dilemma Situations}
	\label{subsec:2.4}
	
	In such instances where the AV predicts that all of its actions will result in a collision, it needs to change its decisional criteria to minimize harm and distribute it in an acceptable way. Thus, even if the reward expression defined above considers a term connected to ethical restrictions, the decision must be taken primarily according to the consequences of the collision. Using the value iteration algorithm (algorithm \ref{alg:1}) to obtain the optimal policy, $\pi^*(s_i)$, (that for each possible state gives the action that should be executed by the AV\footnote{Considering $\pi^*$ as a deterministic policy, which means that each state has an optimal action that should be executed.}), the bifurcation between dilemma and normal situation occurs in the policy optimization operation (for loop in \ref{alg:1}), considering the proposed value iteration algorithm \citep{demoura2020ethical}. Equation \ref{eq:5} continues to be valid, being used in the value maximization step (do-while loop in \ref{alg:1}). 
	
	\begin{center}
		\resizebox{\textwidth}{!}{
			\begin{minipage}{\textwidth}
				\begin{algorithm}[H]
					\DontPrintSemicolon
					\SetKwRepeat{Do}{do}{while}
					\KwData{Spaces S and A, functions R and P, constant $\gamma$, environment data $e$}
					\KwResult{Policy $\pi^*(s)$, Optimal Value $V^{t}(s)$}
					\BlankLine
					$t=0$\;
					\Do{$\tm{MSE}(V^{t+1}-V^{t})>\epsilon$}{
						\For{every $s_i\in S$}{
							\begin{equation*}
								V^{t+1}(s_i) =\max_{a\in A }\left[R(s_i,a,s_{i+1}) + \gamma\cdot\sum_{s_{i+1}\in S'}P(s_{i+1}|s_{i},a)V^{t}(s_{i+1})\right]
							\end{equation*}\;	
							\vspace{-5mm}
							$t=t+1$\;
						}	
					}
					\BlankLine
					\For{every $s_i\in S$}{
						\If{Not a dilemma situation}{
							\vspace{-2.5mm}
							\begin{equation*}
								\pi^*(s_i) = \argmax_{a\in A }V^{t}(s_i)
							\end{equation*}\;
							\vspace{-5mm}
						}
						\Else{
							\vspace{-2.5mm}
							\begin{equation*}
								\pi^*(s_i) = \tm{\textbf{Ethical\_deliberation}}(s_i,A_{\tm{col}},P)
							\end{equation*}\;
							\vspace{-5mm}
						}
					}
					\caption{Value Iteration}
					\label{alg:1}
				\end{algorithm}
			\end{minipage}
		}
	\end{center}
	
	Three parameters are used in the ethical deliberation: the current state $s_i$ (and the set of next states given $s_i$, $S'$), the set of actions $A$ (which in this case is equal to the set of actions that result in a collision, $A_{\tm{col}}$) and the transition probability $P$. In \citep{evans2020ethical} the harm was proposed as the variable to estimate the severity of an accident for each road user (equation \ref{eq:10}). It is defined for a road user $k$ by the multiplication of the difference of velocities before and after the accident ($\vec{v}_f - \prescript{k}{}{\vec{v}}_i$) by the constant of vulnerability ($\prescript{k}{}{c}_{vul}$), which scales the delta of final and initial velocities for a road user to a common severity scale according to the probability of MAIS3+ injury using only the type of the road users involved and the collision configuration. More details are available in \citep{evans2020ethical, moura2021governing}. 
	
	\begin{equation}
		\label{eq:10}
		\prescript{k}{}{h}(s_t,a_t,s_{t+1}) = \prescript{k}{}{c}_{vul}\cdot\left(\| \vec{v}_f - \prescript{k}{}{\vec{v}}_i\|\right)
	\end{equation}
	
	Such quantity is then mixed with the transition probability to obtain the expected harm for a road user $k$:
	
	\begin{equation}
		\label{eq:11}
		\prescript{k}{}{h_{exp}}(s_i,a_j) = \sum_{s_{i+1}\in S'}p(s_{i+1}|s_i,a_j)\prescript{k}{}{h}(s_i,s_{i+1},a_j)
	\end{equation}
	
	Equation \ref{eq:11} gives the final quantity used to deliberate and choose what action the AV should execute given its surroundings.  Using this measure of possible injury severity for the AV and the road users involved and also the ethical valence, some deliberation method (which is the profiles proposed in subsection \ref{subsec:2.1}) must be established to determine which is the best action for the current situation (which were called profiles in section \ref{subsec:2.1}). In \cite{demoura2020ethical}, three approaches, without considering the ethical valence, were proposed: one based on the concept of fairness from \citep{rawls1971theory}, another using the Greatest Happiness Principle from \citep{mill1859utilitarianism} and a last one that tries to address the failure of the utilitarian approach to account for discrepancies on the distribution of happiness \citep{broome2017weighting}. All of them will be modified here to include the valences of each road user in its deliberation.
	
	\subsection*{Contractarian Approaches}
	
	One possible approach to deal with dilemma situations is to define a framework capable of delivering fair choices in such situations. From \cite{rawls1971theory}, two principles of justice are proposed: \emph{"..., each person participating in a practice, or affected by it, has an equal right to the most extensive liberty compatible with a like liberty for all; ..., inequalities are arbitrary unless it is reasonable to expect that they will work out for everyone's advantage ..."}. More specifically, the expression "\emph{must work out for everyone's advantage}" excludes the justification of disadvantages for the worst off so as to give advantages for the better off.
	
	Translating from a generic context to the narrow case of justice in dilemma decision scenarios, one can say that every road user must be awarded the most extensive right to safety that is compatible with the safety of others, and that differences in safety across road users should benefit those who are most at risk. In the imminence of a collision, having the same amount of safety means that ideally the expected harm\footnote{The variable defined in equation \ref{eq:6} which does take into account the differences in nature of road users, not the general idea of risk of harm.} should be equally distributed for all involved road users (involved here refers to all road users engaged in the driving environment), if no other action is available to avoid a collision.
	
	From the previous propositions, three different restrictions can be established (as stated in \cite{moura2021governing}). Firstly, during the deliberation for an action that will inevitably impose the risk of injury to others, the traffic code must continue to be followed. If one considers the traffic code as the mechanism to treat conflicting claims in normal situations, to abandon it in a dilemma situation would open the door to questions about the fairness of an action by the disadvantaged party (which in turn might take the form of legal action against the favored party).
	
	Secondly, road users not involved in the interactions that generated the dilemma situation should not be put in harm's-way, since these road users might not be aware of the risks of such interactions and also because it might be uncertain that they did voluntarily take action to engage in the aforementioned interaction. One simple example can consist of a strict prohibition for the AV to use the sidewalk in any situation, avoiding risking the lives of pedestrians that are not concerned with what happens on the road. The general idea of this restriction is also presented in the guideline 9 proposed by \cite{luetge2017german}.    
	
	These two previous restrictions limit the range of actions that are available for the AV to consider in a dilemma situation. The third one is the translation of the second principle of justice proposed by Rawls: the biggest expected harm should be minimized as long as the expected harms for the other road users do not surpass a danger threshold. The AV's deliberation should try to minimize the largest expected harm (accounting for $w_{\tm{eth}_i}$ as well) while maintaining the others below an acceptable level of risk. Such a danger threshold might be defined statistically based on the harm value that represents a certain probability of severe injury (MAIS3+ level injury \citep{mackenzie1985abbreviated}). However, to determine this threshold it is necessary to use detailed data about the severity of collisions which is not widely available today, apart from private traffic accidentology databases.
	
	Of course, there might be situations where the risk of injury is so high that it is impossible to decrease the expected harm of all the road users (or even one) below the threshold. In this case the ethical valence defined for each road user can determine the greatest need for a lower expected harm, reflecting the hierarchy of strongest claims for safety, as proposed by the definition of ethical valence in subsection \ref{subsec:2.2}.

	\subsection*{Utilitarian Approaches}
	
	This approach is rooted in the Greatest Happiness Principle, which states: "\emph{actions are right in proportion as they tend to promote happiness, wrong as they tend to produce the reverse of happiness. By happiness is intended pleasure, and the absence of pain; by unhappiness, pain, and the privation of pleasure}"  \citep{mill1859utilitarianism}. Happiness, in this definition, refers to the happiness of all people, not only for the individual that is deliberating about an action. 
	
	This idea of the greatest good for everyone embodies the definition of utility, hence every decision should be taken according to the maximization of utility, for the common good. Using the dual formulation of the stated principle, where the pain and unhappiness must be minimized, one should therefore minimize the total amount of expected harm to find the "least wrongful action", as equation \ref{eq:12} shows. The definition of the deliberation rule for this criteria is rather straightforward and involves only the minimization of the sum of expected harms for all road users concerned by the dangerous situation.
	
	\begin{equation}
		\label{eq:12}
		a_{eth} = \argmin_a \sum_{i=0}^{n}\prescript{i}{}{h}_{exp}(a)
	\end{equation}
	
	However, the formulation presented above can also account for the valences if one defines them in the form of weights for each harm quantity, as equation \ref{eq:13} shows:
	
	\begin{equation}
		\label{eq:13}
		a_{eth} = \argmin_a \sum_{i=0}^{n}w_{\tm{eth}_i}\prescript{i}{}{h}_{\tm{exp}}(a)
	\end{equation}
	
	In such a way, depending on the numerical definition of the ethical valences used, the expected harm for a certain road user can be decreased in comparison with another one that has a lower valence. Of course, using the ethical valence in such a way prompts a discussion concerning how to calibrate these weights, which might be done defining a set of standard use cases and the respective desired results to then try to find the range of values for these weights that produce the best acceptable results.
	
	\subsection*{Egalitarian Approaches}
	
	The egalitarian approach addresses one of the most flagrant problems with the utilitarian approach: the lack of constraints concerning extremely unequal distribution of harms between the road users, which in some particular situation can be seen almost as targeting a specific road user to save all others. In the approach called communal egalitarianism presented in \cite{broome2017weighting}, the total sum of goodness delivered by an action can be higher by simply distributing  better the utility between the agents without the need to increase the individual quantities of utility themselves. Using the variance of the action being evaluated for the road user i, $\tm{var}_{h_{\tm{exp}}}(i,a)$, as a measure of disparity between actions and the ethical valence as weight, equation \ref{eq:14} determines the best action given the discussed paradigm: 
	
	\begin{equation}
		\label{eq:14}
		a_{eth} = \argmin_a \sum_{i=0}^{n}w_{\tm{eth}_i}\cdot \tm{var}_{h_{\tm{exp}}}(i,a)\cdot\prescript{i}{}{h}_{\tm{exp}}(a)
	\end{equation}
	
	Since the variance was used to represent the disparity between actions, it implements a quadratic relationship restriction on the possible inequalities of distribution. Of course, another measure could be chosen, for example the absolute value of the difference between the mean expected harm and the one produced by the action in question ($\prescript{i}{}{h}_{\tm{exp}}(a)$) for the road user $i$. 
	
	\section{Conclusion}
	\label{sec:4}
	
	The advent of advanced forms of automated systems, such as automated vehicles, has undeniably brought these machines into situations of ethical salience. As we have argued, the fact that these machines are quickly becoming the subjects of human moral judgement when acting in specific contexts does not imply that these machines either are, or ought to be endowed with a corresponding anthropomorphic capacity for moral judgement. At base, beyond a purely syntactic processing of contextual data, spontaneously acting machines remain ignorant of the conceptual and normative realms. Nevertheless, these machines must still be equipped with a tactical form of ethical sensitivity, one which seeks to minimize or neutralize the harm they are liable to inflict on the human agents with whom they interact. The Ethical Valence Theory is an attempt to formalize this tactical approach to outwardly ethical behavior in machines. Importantly, if machines are structurally immune to the perplexities that spring from the plurality of moral principles and methods which inform human ethics, the designers of these machines are not. In this sense, moral paradigms such as contractualism, utilitarianism or egalitarianism constitute different robust approaches to the mitigation of the harm machines may cause, whose algorithmic implementation is ultimately less problematic than their selection and eventual justification in the eyes of society. Thus, a deflationist and tactical approach to the ethics of automated vehicles shifts the locus of an AV’s moral agency to its rightful source: the actions and decisions of the human designers, manufacturers, and stakeholders responsible for the AV’s design, development and deployment. 
	
	
	\bibliography{book_refs.bib}
	\bibliographystyle{apacite}
	
\end{document}